\documentclass[graybox]{svmult}

\usepackage{mathptmx}       
\usepackage{helvet}         
\usepackage{courier}        
\usepackage{type1cm}        
\usepackage{makeidx}         
\usepackage{graphicx}        
\usepackage{multicol}        
\usepackage[bottom]{footmisc}

\makeindex             

\begin{document}

\title*{Energy cost reduction in the synchronization of a pair of nonidentical coupled Hindmarsh-Rose neurons}
\titlerunning{Energy cost reduction in the synchronization of nonidentical coupled neurons }
\author{A. Moujahid, A. D'Anjou, F.J. Torrealdea and C. Sarasola}
\institute{A. Moujahid \at Computational Intelligence Group, Department of Computer Science, University of the Basque Country UPV/EHU, Spain, \email{jibmomoa@gmail.com}
\and A. D'Anjou \at Computational Intelligence Group, Department of Computer Science, University of the Basque Country UPV/EHU, Spain
\and F.J. Torrealdea \at Computational Intelligence Group, Department of Computer Science, University of the Basque Country UPV/EHU, Spain
\and C. Sarasola \at Department of Physics of Materials, University of the Basque Country UPV/EHU, Spain}
%
%
\maketitle

\abstract{Many biological processes involve synchronization between nonequivalent systems, i.e, systems where the difference is limited to a rather small parameter mismatch. The maintenance of the synchronized regime in this cases is energetically costly \cite{1}. This work studies the energy implications of synchronization phenomena in a pair of structurally flexible coupled neurons that interact through electrical coupling. We show that the forced synchronization between two nonidentical neurons creates appropriate conditions for an efficient actuation of adaptive laws able to make the neurons structurally approach their behaviours in order to decrease the flow of energy required to maintain the synchronization regime.}

\section{Introduction}
\label{sec:1}

When a given oscillator moves
freely on its natural attractor its oscillatory regime consists of
a balanced exchange of energy between the system and its
environment that occurs spontaneously through the divergent
components of the system’s structure without concurrence of
any additional device. If, on the other hand, the system is
forced to synchronize to a different guiding system its oscillatory
regime occurs on an unnatural region of the state space
where there is a nonzero net average exchange of energy
with its environment. This net flow of energy per unit time
requires the concurrence of a coupling device that includes
an external source of energy. This flow of energy is necessary
to maintain the synchronized regime and constitutes a
cost for the synchronization process \cite{1}. This consumption of energy can be reduced if the guided
system itself adapts its structure to become closer to the one
of the guiding system \cite{3}. Ideally, if the systems become identical
their joint dynamics is attracted toward a regime of zero
error in the variables.

Many biological processes involve synchronization between different members of the same family of systems that have similar, although not identical, values of some distinctive parameters. This work studies the energy implications of synchronization phenomena in a pair of structurally flexible coupled neurons that interact through electrical coupling. We show that the forced synchronization between two nonidentical neurons creates appropriate conditions for an efficient actuation of adaptive laws able to make the neurons structurally approach each other in order to decrease the flow of energy required to maintain the synchronization regime. The neuron has been modelled by a four-dimensional Hindmarsh-Rose model \cite{6,7,8,9,10}.
This model is described by the following equations of movement:

\begin{equation}
\begin{array}{l}
            \dot{x}=ay+bx^2-cx^3-dz+\xi I,    \\
            \dot{y}=e-fx^2-y-gw,      \\
            \dot{z}=m(-z+s(x+h)), \\
 	    \dot{w}=n(-kw+r(y+1)),
\end{array}
\label{eq1}
\end{equation}
\begin{figure}[ht]
\begin{center}
\includegraphics [width=10cm,height=6cm]{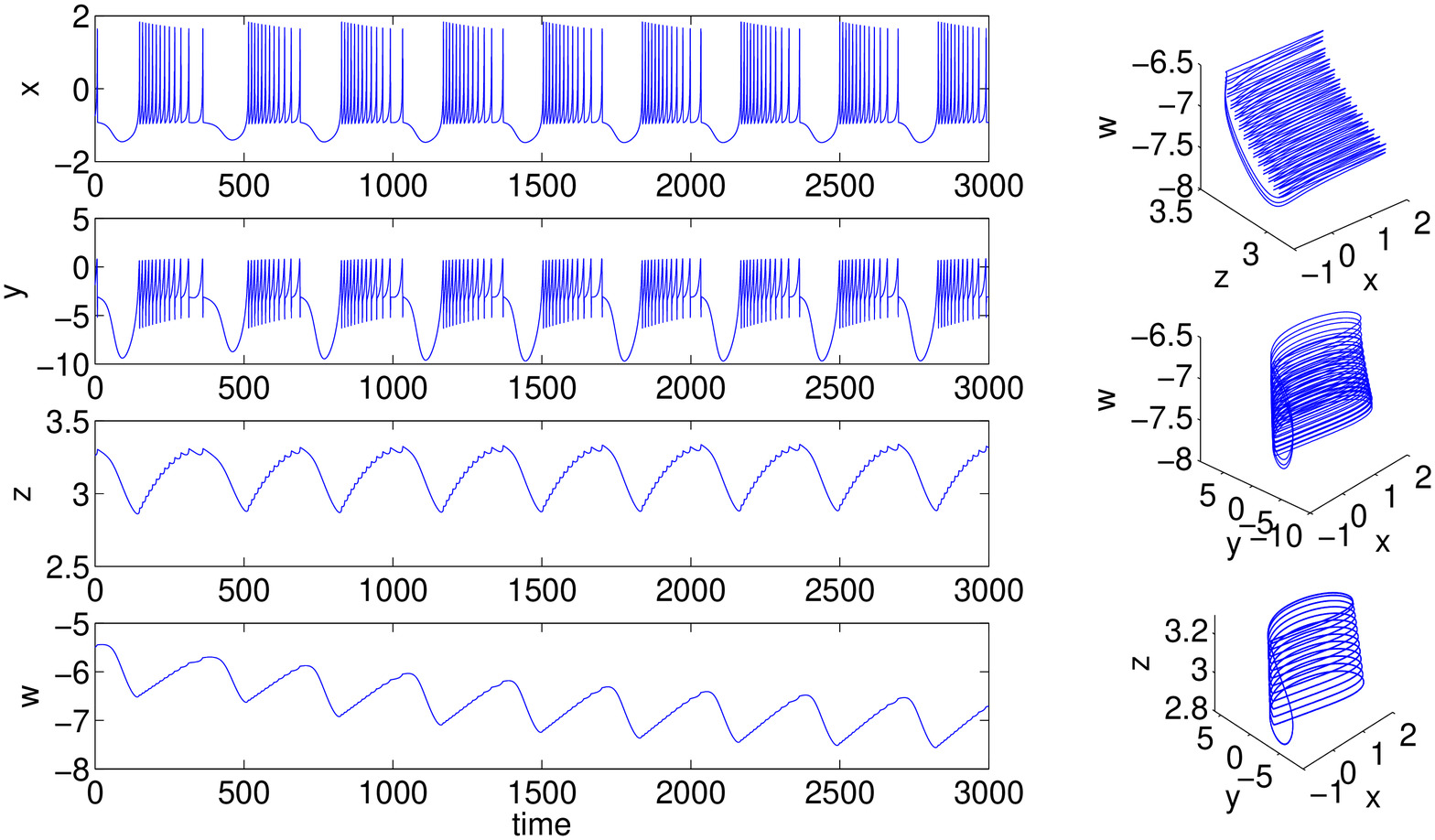} \end{center}
\caption{Time series and 3D projections of the dynamical variables $x(t)$, $y(t)$, $z(t)$, $w(t)$ of the four-dimensional Hindmarsh-Rose neuron model}
\label{f1}
\end{figure}
where $a,b,c,d,\xi,I,e,f,g,m,s,h,n,k,r$, and $l$ are the parameters that govern the dynamics of the neural system. The variable $x$ is a voltage associated to the membrane potential, variable $y$ although in principle associated to a
recovery current of fast ions has been transformed into a voltage, variable $z$ is a slow adaptation current associated to
slow ions, and variable $w$ represents an even slower process than variable $z$ \cite{11}. $I$ is a external current input, and is the main parameter we used to control the modes of spiking and bursting activity of the model.
For the numerical results of this work we fix the parameters to the values
$a = 1$, $b = 3.0 (mV)^{-1}$, $c = 1 (mV)^{-2}$, $d = 0.99M\Omega$, $\xi = 1M\Omega$,
$e = 1.01mV$, $f = 5.0128 (mV)^{-1}$, $g = 0.0278M\Omega$, $m = 0.00215$, $s = 3.966\mu S$, $h = 1.605 mV$, $n = 0.0009$, $k = 0.9573$, $r = 3.0\mu S$, $l = 1.619mV$.

Figure 1 shows a chaotic time series of the four variables. The complexity achieved by the incorporation of a slow variable $w$ that increases the realism of the description of slow Calcium currents can be observed in the projections of the attractor on the ($x,y,z$), ($x,y,w$) and ($x,z,w$) axes.

In Sec. 2 we report the energy-like function associated to a four-dimensional Hindmarsh-Rose model. Sec. 3 briefly summarizes the adaptation mechanism we used to adapt the structure of the postsynaptic neuron, and presents computational results of the synchronization process of two electrically nonidentical coupled neurons. We consider that the presynaptic (sending) neuron always signal in  a chaotic regime, while the postsynaptic (receiving) neuron is set to its quiescent state at a low value of its external current.  In a first stage, the postsynaptic neuron has been forced to synchronize with the presynaptic one, then we initiate an adaptive process that adapts some parameters of the postsynaptic neuron to ones of the presynaptic neuron. We have analysed the energy dissipation of the receiving neuron during the synchronization process without and with structural adaptation.

\section{Four-dimentional Hindmarsh-Rose model energy}

\begin{figure}[ht]
\begin{center}
\includegraphics [width=10cm,height=6cm]{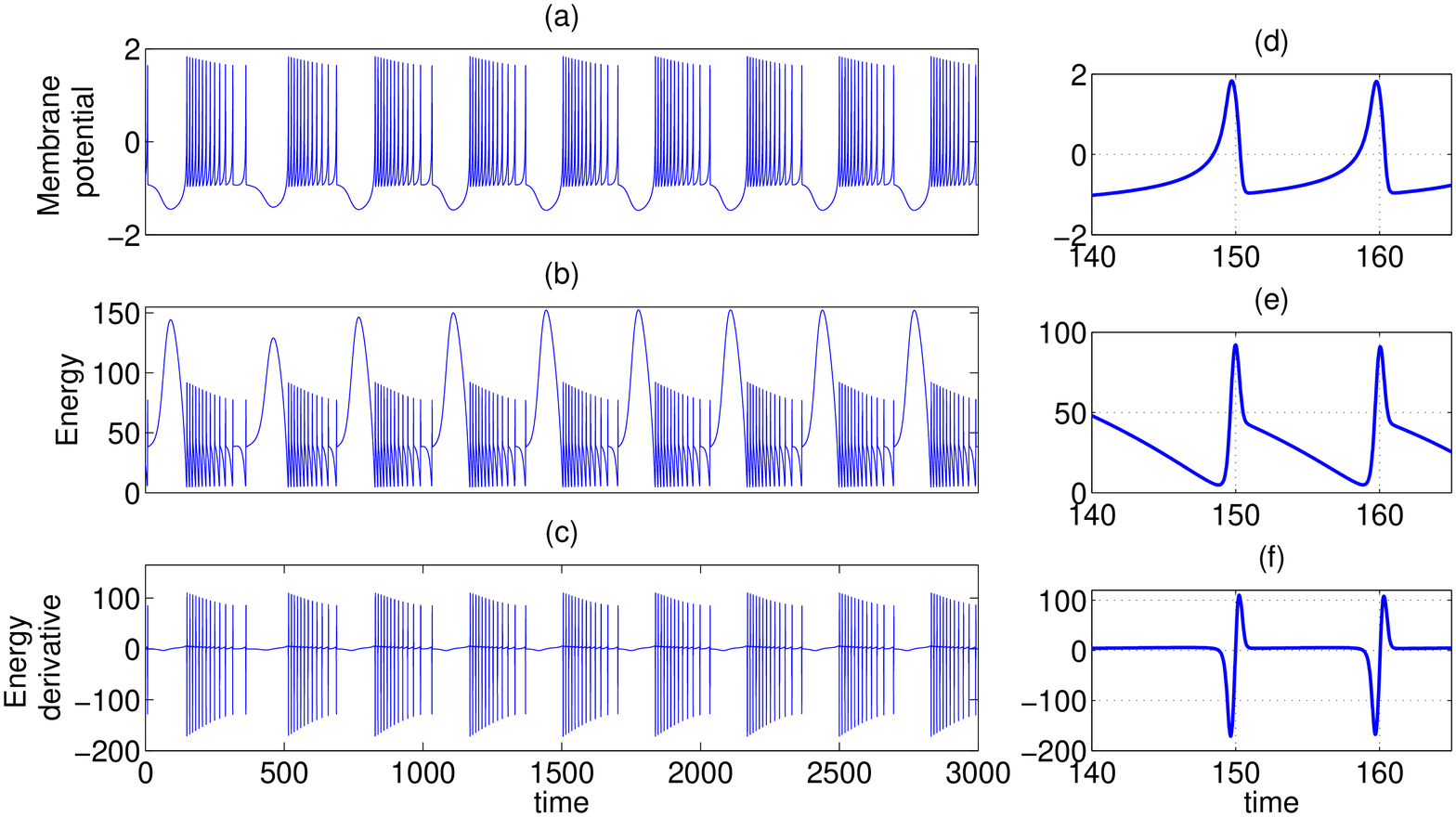}
\caption{(a) Action potentials, (b) energy and (c) energy derivative for the Hindmarsh-Rose model neuron. (d), (e) and (f) Details of the action potential, energy and energy derivative associated to two spikes}
\end{center}
\label{f2}
\end{figure}

In the Hindmarsh-model given by Eq. (1) the energy function $H(x)$ is given by \cite{2}
\begin{equation}
 H=\frac{p}{a}(\frac{2}{3}fx^3+\frac{msd-gnr}{a}x^2+ay^2) \\
+\frac{p}{a}(\frac{d}{ams}(msd-gnr)z^2-2dyz+2gxw)
\label{eqq2}
\end{equation}
where p is a parameter. As in the model time is dimensionless and
every adding term in Eq.(\ref{eqq2}) has dimensions of square voltage,
function H is dimensionally consistent with a physical energy as
long as parameter p has dimensions of conductance. In this paper
we fix parameter p to the arbitrary value $p = -1 S$. The minus sign
has been chosen to make consistent the outcome of the model with
the usual assumption of a demand of energy associated with the
repolarization period of the membrane potential and also with its
refractory period (see Fig. \ref{f2}).

And the corresponding energy derivative $\dot{H}$ is given by \cite{2}
\begin{equation}
\dot{H}= \frac{2p}{a}\left( \begin{array}{c}
            fx^2+\frac{msd-gnr}{a}x+gw     \\
            ay-dz      \\
            \frac{d}{ams}(msd-gnr)z-dy \\
 	    gx
           \end{array}
    \right)
\left( \begin{array}{c}
            bx^2-cx^3+\xi I     \\
            e-y      \\
            msh-mz \\
 	    nrl-nkw
           \end{array}
    \right)
\end{equation}
is also dimensionally consistent with a dissipation of energy. As
the states of an isolated Hindmarsh–Rose neuron are confined to
an attractive manifold the range of possible values of its energy is
recurrent and the long term average of its energy derivative is zero.

This energy and energy derivative functions are used to evaluate the energy consumption of the neuron in isolation and also when it is connected
to other neurons through electrical synapses, and provide the basis for all the computational results presented in this work. The procedure followed to find this energy function has been reported in detail in \cite{1}.

Figure \ref{f2} (a) shows a series of action potentials (variable $x$ in the model neuron). Fig. \ref{f2}(b) and Fig. \ref{f2}(c) show both energy and energy derivative corresponding to that action potentials. Fig. \ref{f2}(e) and Fig. \ref{f2}(f) show  detail of energy and energy derivative associated to a train of two action potentials.
For each action potential it can be appreciated (see Fig. \ref{f2}(e,f)) that the energy derivative
is first negative, dissipation of energy while the membrane
potential depolarizes during the rising period of the spike, and
then positive, contribution of energy to repolarize the membrane
potential during its descending period. During the refractory period
between the two spikes the energy derivative remains slightly positive,
still demanding energy, until the onset of the following action
potential.

\section{Synchronization energy of two electrically coupled neurons}

In this section we analyze the energy aspects of the synchronization of two nonidentical neurons coupled by an electrical synapse. The presynaptic neuron is set in the chaotic spiking-bursting regime corresponding to an external current $I_1=3.024$. While the postsynaptic neuron is set to its quiescent state at a low value $I_2=0.85$ of its external current. The two neurons are coupled unidirectionally according to the following equations:

\begin{equation}
\begin{array}{l}
            \dot{x}_i=ay_i+bx_i^2-cx_i^3-dz_i+\xi I_i+K_i(x_j-x_i), \\
            \dot{y}_i=e-fx_i^2-y_i-gw_i,      \\
            \dot{z}_i=m(-z_i+s(x_i+h)), \\
 	    \dot{w}_i=n(-kw_i+r(y_i+1)),
\end{array}
\label{eq2}
\end{equation}
where $K_1=0$ and $K_2 \geq 0$ is the coupling strength. $i,j=1,2;i\neq j$ are the indices for the neurons. Note that the coupling affects only the first variables $x_2$ of the postsynaptic neuron.

The coupling scheme given in Eq. \ref{eq2} can be written in general terms as follows:

 \begin{equation}
\begin{array}{l}
            \dot{y}=f(y,p) \\
            \dot{x}_k=f(x_k,q)+K(y-x_k),      \\
\end{array}
\label{eq3}
\end{equation}
where $p$ and $q$ stand for the parameters of the presynaptic and postsynaptic neurons, and $x_k(t)$ indicates the states of the postsynaptic neuron when the coupling strength is set to $k$.  Notice that $K(y-x_k)$ is the coupling interface required in order to be physically able to implement the coupling of both neurons $\dot{x}=f(x)$ and $\dot{y}=f(y)$.

If the coupling strength $K$ is large enough as to make the errors in the variables $e=x_k-y$ small, an operational law that adapts the parameters of the postsynaptic neuron to the ones of the presynaptic neuron is given by \cite{3}

 \begin{equation}
\begin{array}{l}
            \dot{e}_i^p=- \left[ \sum_{l=1}^n \left( \frac{\partial f_l(x_k,q)}{\partial q_i} \right) _{(y,p)} e_l \right] \\
\end{array}
\label{eq4}
\end{equation}
where $e^p=q-p$ denotes the vector of parameter errors, and the summation is over every component of the vector field $f$. The above law is general
and can be used to find specific adaptive laws to any kind of homochaotic systems provided they are coupled through a
feedback scheme of large enough coupling strength.

\begin{figure}[ht]
\begin{center}
\includegraphics [width=12cm,height=4cm]{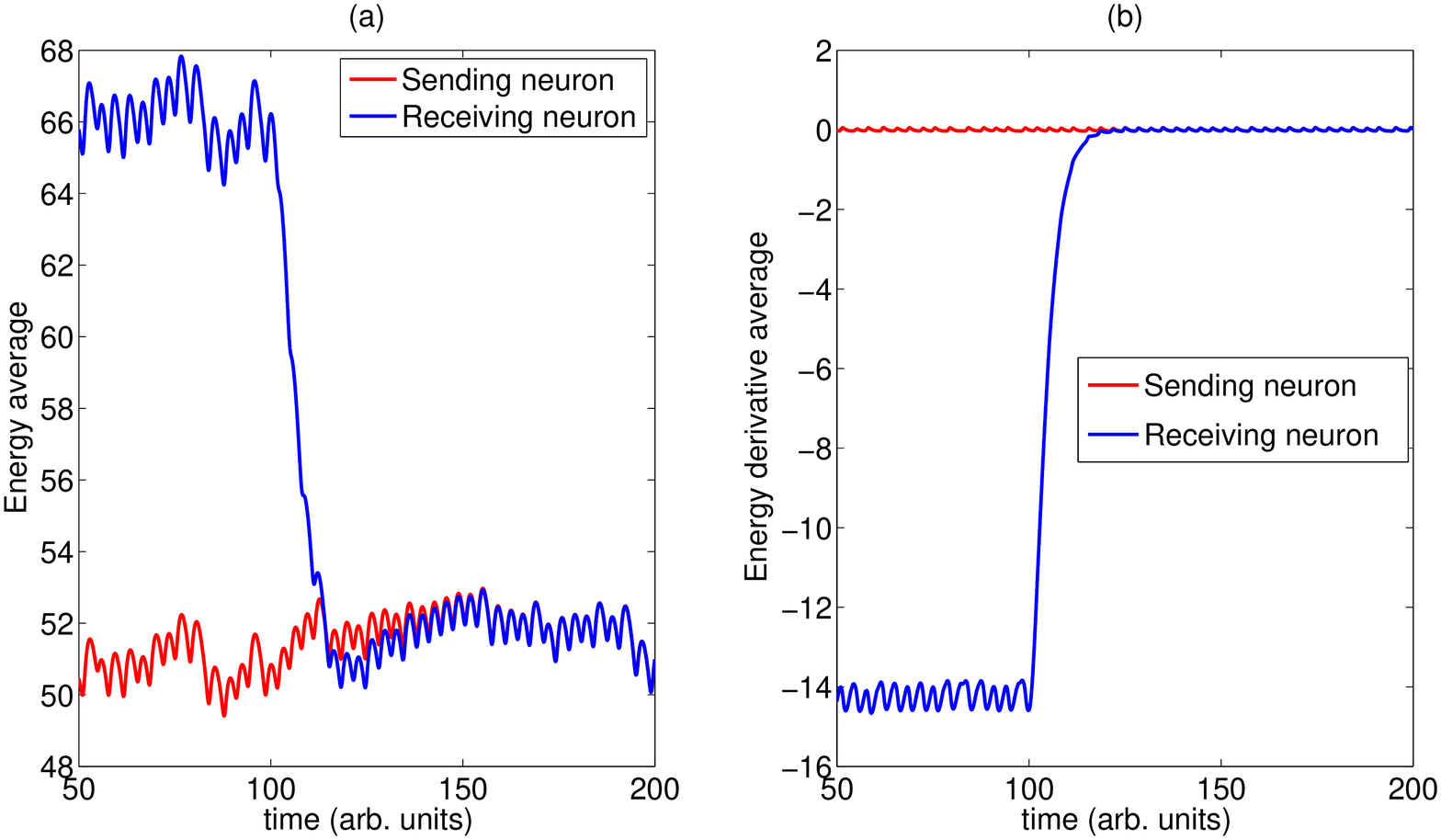} \end{center}
\caption{(a) Average over ten units of time of the energy per unit time. (b) Average over five units of time of the energy derivative per unit time. Adaptation of the external current $I_2$ begin at $t=100$. The coupling strength is set to value $k=5$}
\label{f3}
\end{figure}

In the following we analyse the change in the balance of energy of the postsynaptic neuron when its external current parameter is governed by an adaptive law in order to reach the nominal value of the external current in the presynaptic neuron. The adaptation law has been implemented following Eq. \ref{eq4}. For this experiment, we have used a coupling strength with value $k=5$. We started the adaptation procedure at $t=100$ and registered data between $t=50$ and $200$ for proper observation of the evolution of both energy and energy derivative during the process. The registrated values has been averaged over a convenient length of time in order to avoid large fluctuations. The dissipated energy (energy derivative) has been averaged over five units of time, while the proper energy has been averaged over ten units of time.

Figure \ref{f3} shows the average values of both energy and energy derivative per unit time. In Fig. \ref{f3} (a) we can see that in a first stage ($t<100$), the receiving neuron is forced to synchronize with the sending neuron, and oscillates in an unnatural region of the state space characterized by an average energy of about 66 (arb. units), and a nonzero energy derivative average of about 14 (arb. units), ie,  a nonzero net average exchange of energy with its environment (see Fig \ref{f3} (b)). After adaptation take place, the two neurons become structurally close each other, and enter in a completely synchronized regime of balanced exchange of energy between the system and its environment corresponding to zero value of energy derivative ($\dot{H}=0$).

\begin{figure}[ht]
\begin{center}
\includegraphics [width=10cm,height=6cm]{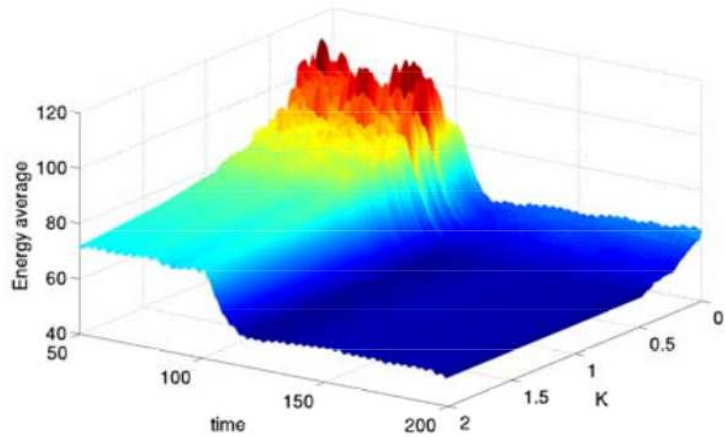} \end{center}
\caption{Average over five units of time of energy per unit time  of the receiving neuron at different values of the coupling strength $K$. Adaptation of the external current $I_2$ begin at $t=100$}
\label{f4}
\end{figure}

To illustrate  the ability of this adaptive laws to decrease the energy dissipation, we have computed, for the receiving neuron, the average energy and average energy dissipation over a convenient length of time at different values of the coupling strength ranging from $K=0$ to $K=2$. The adaptation run at time $t=100$.

Figure \ref{f4} shows average over ten units of time of the energy per unit time at different values of the coupling strength ranging from $k=0$ to $k=2$. It can be seen that before the adaptation occurs ($t<100$), and for low value of the coupling strength not sufficient to induce a certain degree of synchrony, the receiving neuron shows a waving average energy pattern. This oscillating regime of energy average disappears once the adaptation process start, and decreases to low values. For high  values of the coupling strength, the receiving neuron is forced to synchronize with the sending neuron and moves in a region of state space where the average of its energy is greater than that of the sending neuron. When the coupling strength is large enough as to make the errors in the variables small, and as soon as the adaptation process starts the average energy quickly decreases to values corresponding to the average energy of the sending neuron.

\begin{figure}[ht]
\begin{center}
\includegraphics [width=12cm,height=7cm]{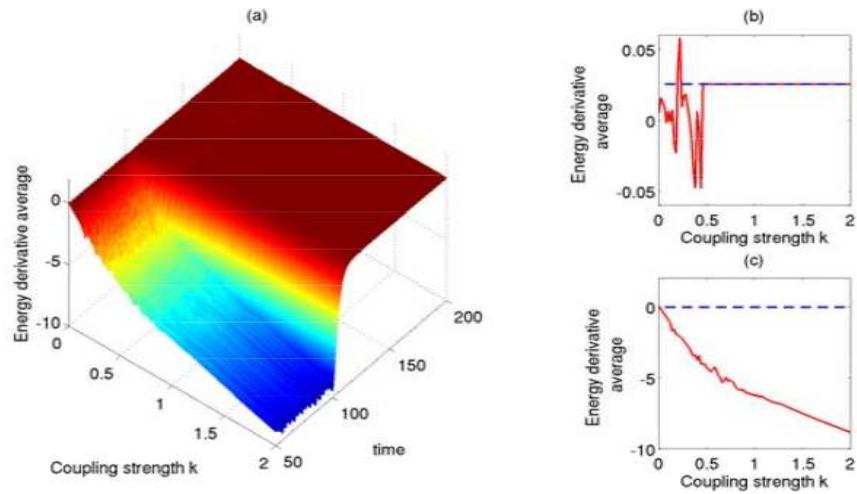} \end{center}
\caption{(a) Average over ten units of time of of energy derivative per unit time of the receiving neuron at different values of the coupling strength $K$. (b) and (c) show respectively the average of energy derivative of the receiving neuron (solid line) and the sending neuron (dash line) as a function of the coupling strength with and without structural adaptation. Adaptation of the external current $I_2$ begin at $t=100$.  }
\label{f5}
\end{figure}

Figure \ref{f5} shows average over five units of time of the energy derivative per unit time at different values of the coupling strength ranging from $k=0$ to $k=2$. When $K=0$, ie, no guidance at all, the receiving neuron moves on its natural region of state space and its averaged dissipated energy is zero. As soon as the coupling device is connected the average energy derivative per unit time becomes negative, that is, it start to dissipate on average an energy that the coupling device will have to provide in order to maintain the forced regime. The required energy increases with coupling strength as it can appreciated in Fig.\ref{f5} (c). Once the adaptation process starts the average energy derivative quickly decreases to zero, reflecting the fact the receiving neuron has become structurally so close to the sending neuron that they can reach a regime of identical synchronization.

%
%

\end{document}